\documentclass{CVIS}
\usepackage{amsmath,bm}
\usepackage{graphicx}
\usepackage{dcolumn}
\usepackage{subcaption,soul}
\usepackage{caption}
\usepackage[numbers,sort&compress,sectionbib]{natbib}
\usepackage[pagebackref=flase,breaklinks=true,letterpaper=true,colorlinks=true,linkcolor = blue, urlcolor = black,citecolor=black,bookmarks=true]{hyperref}
\usepackage{url}
\usepackage{mleftright}
\usepackage{multirow}

\usepackage{titlesec}
\usepackage[redeflists]{IEEEtrantools}

\title{Causal Discovery from Sparse Time-Series Data Using Echo State Network}

\begin{document}
\bstctlcite{IEEEexample:BSTcontrol}
\sloppy

\author{
\begin{tabularx}{\textwidth}{X X}
Haonan Chen$^{\dag1}$ & University of Waterloo, Waterloo, ON Canada\\
Bo Yuan Chang$^{\dag1}$  & University of Waterloo, Waterloo, ON Canada\\
Mohamed A. Naiel$^1$ & University of Waterloo, Waterloo, ON Canada\\
Georges Younes$^1$ & University of Waterloo, Waterloo, ON Canada\\
Steven Wardell$^2$ & ATS Automation, Cambridge, ON Canada\\
Stan Kleinikkink$^2$ & ATS Automation, Cambridge, ON Canada\\
John S. Zelek$^1$ & University of Waterloo, Waterloo, ON Canada\\
\end{tabularx}
Email: $^1$\{haonan.chen, by2chang, mohamed.naiel,georges.younes,jzelek\}@uwaterloo.ca
\\\qquad\quad$^2$\{swardell, skleinikkink\}@atsautomation.com
}
\maketitle
\begin{abstract}
Causal discovery between collections of  time-series data can help diagnose causes of symptoms and hopefully prevent faults before they occur. However, reliable causal discovery can be very challenging, especially when the data acquisition rate varies (\textit{i.e.,} non-uniform data sampling), or 
in the presence of missing data points (\textit{e.g.,} sparse data sampling). 
To address these issues, we propose a new system comprised of two parts, the first part fills missing data with a Gaussian Process Regression, and the second part leverages an Echo State Network, which is a type of reservoir computer (i.e., used for chaotic system modelling) for Causal discovery.

We evaluate the performance of our proposed system against three other off-the-shelf causal discovery algorithms, namely, structural expectation maximization, sub-sampled linear auto-regression absolute coefficients, and multivariate Granger Causality with vector auto-regressive using the Tennessee Eastman chemical dataset; we report on their corresponding Matthews Correlation Coefficient (MCC) and Receiver Operating Characteristic curves (ROC) and show that the proposed system outperforms existing algorithms, demonstrating the viability of our approach to discover causal relationships in a complex system with missing entries. 
\end{abstract}
\footnote{${^{\dag}}$ Equal contribution and work was done at Vision and Image Processing Group.}

\section{Introduction}
Artificially replicating and surpassing human's ability to interpret causal relationships is considered an evolutional step forward when compared to  existing Artificial Intelligence  systems \cite{Judea_Peral_causal_theory}.
Despite several decades of research, current state-of-the-art deep learning systems are still broadly interpreted as a family of highly nonlinear statistical models \cite{barlett_2021_arxiv} or pattern detection regression engines that require a very large set of training samples to distill a meaningful outcome for a particular problem. In contrast, humans are capable of drawing conclusions and solving problems by identifying causal relationships between the different variables of a task using a very small amount of data.
This inspired a large body of researchers to work on causal discovery algorithms in an attempt to answer various challenges in several fields, such as epidemiology \cite{causal_HIV}, economy \cite{causal_economy,causal_economy_1}, and medicine \cite{causal_medicine} to name a few.
However, due to the complex relationships between a large number of hidden variables, learning causal relationships on real-world data can be very challenging as causality must be inferred from noisy data  \cite{survey_causal_2020,survey_causal_2020_1}.
Moreover, it is not uncommon to run into missing values when working with real-world data, especially when hardware limitations and human errors play an important factor in the data acquisition process (\textit{e.g.,} time series data). 

Causal discovery in time-series data is of particular importance in this work as we attempt to retrieve causal relationships from sparsely sampled time-series data.  Despite the existence of several methods for causal discovery \cite{slarac,ABCNN_causal,mvgc}, there is currently a limited availability for methods that address causal discovery from \textit{sparse} time-series data.
While the effects of missing data points from a sparsely sampled sequence can be often solved with hardware upgrades, or better training for the employees, such solutions are often associated with a steep price tag and can be time consuming. Alternatively, one could make use of data filling interpolation methods to populate the sparse time-series sequence before performing causal learning as in \cite{Bo_CVIS}. To that end, we propose a system that makes use of Gaussian Process Regression (GPR) to fill missing data points, which are in turn processed with an Echo State network tailored to the causal discovery of relationships between its nodes.

\section{Architecture}
\begin{figure}[!htb]
    \centering
    \includegraphics[width=0.5\textwidth,height=0.5\textheight,keepaspectratio]{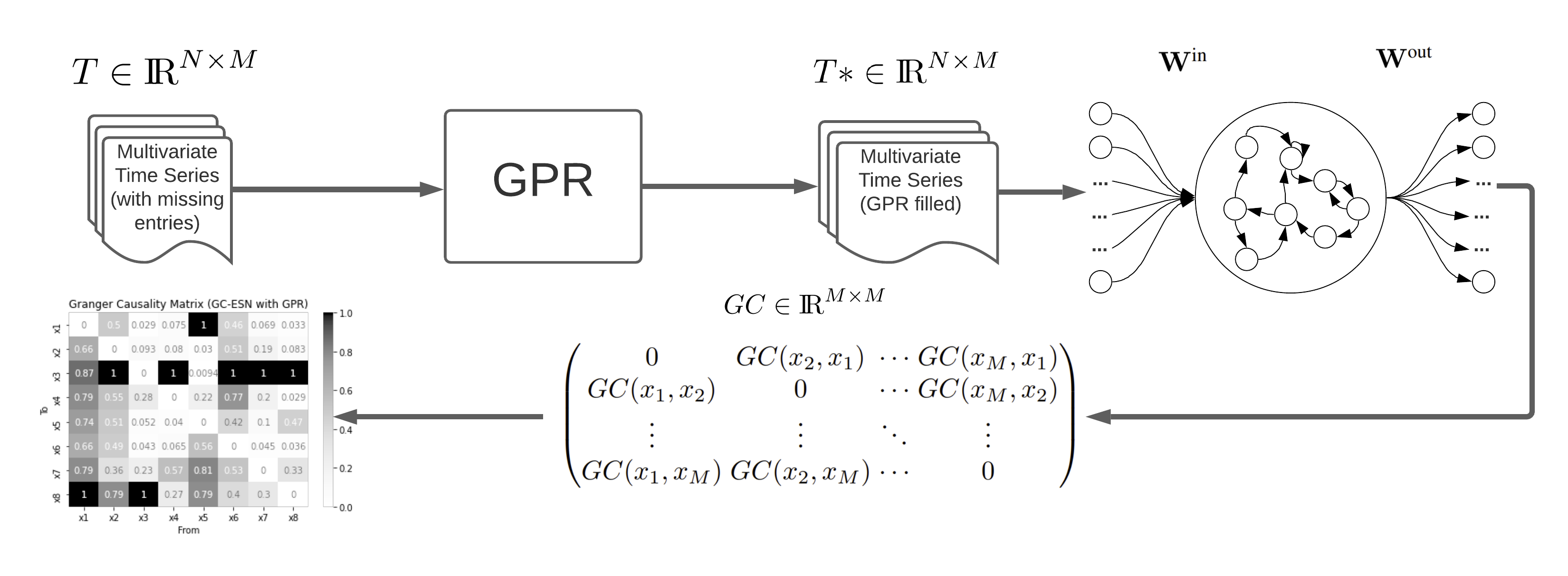}
    \caption{Workflow of the proposed solution; a time-series with missing data is filled using GPR and fed into a GC-ESN to predict causal relations.}
    \label{fig:2}
\end{figure}

Our proposed approach is summarized in Fig. \ref{fig:2}
where $T$ is a time-series input with some missing data, $T^{*}$ represents the multivariate time-series that was data-filled by GPR, and $GC(x_i,x_j)$ is the causal relation between $x_i$ and $x_j$. The input is a $N \times M$ multivariate time-series where $N$ is the total number of entries for each variable (feature) and $M$ is the number of variables. The GPR filled time-series is then processed through 
the GC-ESN estimator (dynamic reservoir), to generate an $M\times M$ causality matrix, which can then be displayed as a heat map of causal relations amongst variables. 

\section{Experiments and Evaluation}
\subsection{Dataset Description}
We evaluate our proposed system on the Tennessee Eastman's Process Dataset (or TE for short) \cite{TE_Dataset}. TE is a dataset that simulates actual chemical processes, and has been widely applied 
in the study of fault diagnosis and root cause analysis \cite{Bayesian_Causal_Analysis_TE} \cite{Root_Cause_KPCA_TE}. The process flow consists of 5 major physical components: the reactor, condenser, vapor-liquid separator, compressor, and the product stripper. There are several sensor measurements available (such as flow rate, temperature, pressure and feed rate) for each components. Overall, the TE dataset contains 500 operation cycles' measurement from 52 sensor readings. Each operation cycle consists of 500 entries being recorded every 3 minutes for a total duration of 1500 minutes. For further information about the TE dataset, the interested reader is referred to \cite{TE_Data_Implementation_Detail}.

We select 8 sensors readings from the first 200 operation cycle's entry inside the training fault free file for our experiment. The description for the 8 selected variables are shown in Table \ref{tab:Feature Description}. In the the experimental procedure, we aim to recover these causal relationships and compare against their ground truth values.
\begin{table}[!htb]
    \caption{Selected Sensor's Descriptions}
    \centering
    \resizebox{\columnwidth}{!}{
\begin{tabular}{|c|c|c|c|}
\hline
Variable ID & Header in data & Description                              & Units                           \\ \hline
1           & xmeas\_5       & Recycle Flow                             & km$^3$/h         \\ \hline
2           & xmeas\_6       & Reactor feed rate                        & km$^3$/h         \\ \hline
3           & xmeas\_7       & Reactor Pressure                         & kPa                             \\ \hline
4           & xmeas\_8       & Reactor Level                            & \%                              \\ \hline
5           & xmeas\_9       & Reactor Temperature                      & \textdegree{}C \\ \hline
6           & xmeas\_12      & Separator Level                          & \%                              \\ \hline
7           & xmeas\_20      & Compress Work                            & KW                              \\ \hline
8           & xmeas\_21      & Reactor cooling water Outlet Temperature & \textdegree{}C \\ \hline
\end{tabular}
}
\label{tab:Feature Description}
\end{table}

\subsection{Experimental Setup}
\begin{figure}[!hbt]
    \centering
    \includegraphics[width=\columnwidth,height=\textheight,keepaspectratio]{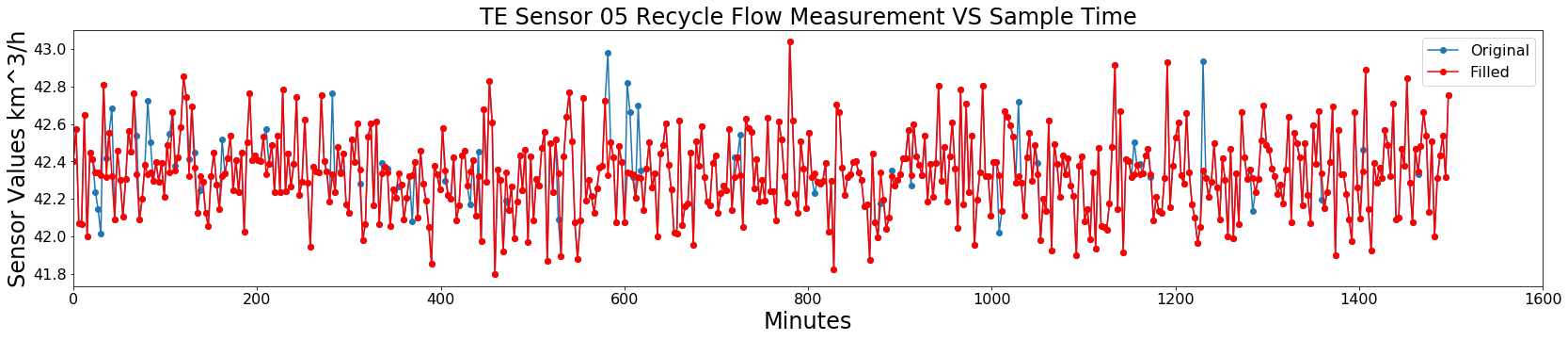}
    \caption{Original Data VS 10\% GPR Filled Data}
    \label{fig:5}
\end{figure}

The selected TE data is first contaminated by removing 10\% of its original entries. This provides ground truth data to validate the success of our proposed data filling process.
Figure \ref{fig:5} shows the original \textit{vs.} filled data for sensor 5 (recycling flow rate measurement).\\
The contaminated data is then processed through the proposed GPR method to fill out the missing entire, and subsequently fed into the 
Echo State Network to discover causal relationships among the variables.

\begin{figure}[!htb]
    \centering
    \includegraphics[width=0.5\textwidth,height=0.3\textheight,keepaspectratio]{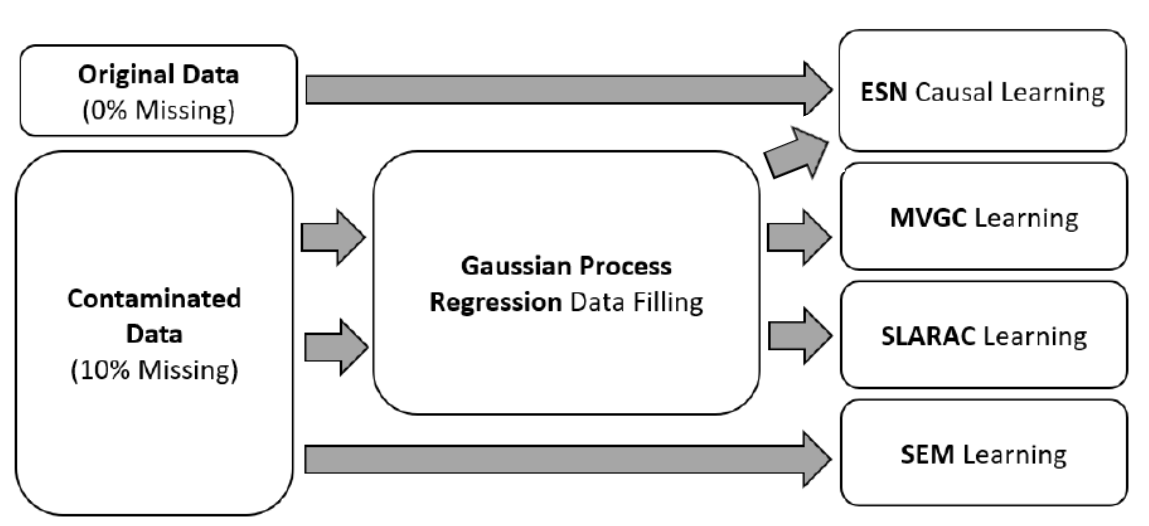}
    \caption{The Proposed Experiment Setup for Performance Comparison.}
    \label{fig:6}
\end{figure}

\noindent{The} process is repeated on three other 
off-the-shelf causal discovery algorithms, namely:
\begin{itemize}
    \item Structural Expectation Maximization (SEM) \cite{SEM}: a graphical approach that is capable of learning causal relationships from sparsely sampled time series data; as such, the GPR process is not used for this algorithm, and the missing data are directly fed into the SEM algorithm for causal learning. We report on the results of \textit{bnstruct} \cite{bnstruct}, an R implementation of SEM.
    \item Subsampled Linear Auto-Regression Absolute Coefficients algorithm (SLARAC): a classical approach for causal discovery written in python, and can be found at \cite{slarac}.
    \item Multivariate Granger Causality (MVGC): a classical approach written in Matlab and can be found at \cite{mvgc}.
\end{itemize}
\noindent{The} GPR filling is applied to both classical approaches before performing causal discovery. To validate the impact of the data filling process, we also include our results obtained from then ESN with the original data (no missing values).\\ 
The experimental comparison setup for the various systems is summarized in Fig. \ref{fig:6}. 


\subsection{Evaluation Metrics}


We validate the performance of the various systems using the Matthews Correlation Coefficient (MCC) score:
\begin{align}
\textrm {MCC} = \frac{TP\times TN - FP\times FN}{\sqrt{(TP+FP)(TP+FN)(TN+FP)(TN+FN)}},    
\label{eq:mcc}
\end{align}
that reflects the \textit{usefulness} of each causal prediction. The MCC score provides a balanced measure between the True Positive (TP), True Negative (TN), False Positive (FP) and False Negative (FN) cases; its output is $\in [-1,1]$, where a $+1$ score indicates a perfect prediction, $0$ indicates that the prediction made is no better than random guessing, and $-1$ indicates complete disagreement between prediction and observation \cite{MCC}.\\

In addition to the MCC index comparison, we also report on the ROC (Receiver Operating Characteristic) curves for the various causal discovery systems. The ROC curves offer valuable insights into the specificity and sensitivity of each model at different cutoff thresholds in the causal matrices.
We also report on the AUC (Area Under Curve) score for all ROC curves.
Note that due to the binary matrix output nature of the SEM algorithm, we were unable to plot its ROC curve.



\subsection{Results and Discussion}
\begin{figure}[htp]
\centering
\subfloat[Causal Diagram of GC-ESN with 10\% GPR]{%
  \includegraphics[clip,width=0.7\columnwidth,height=6cm]{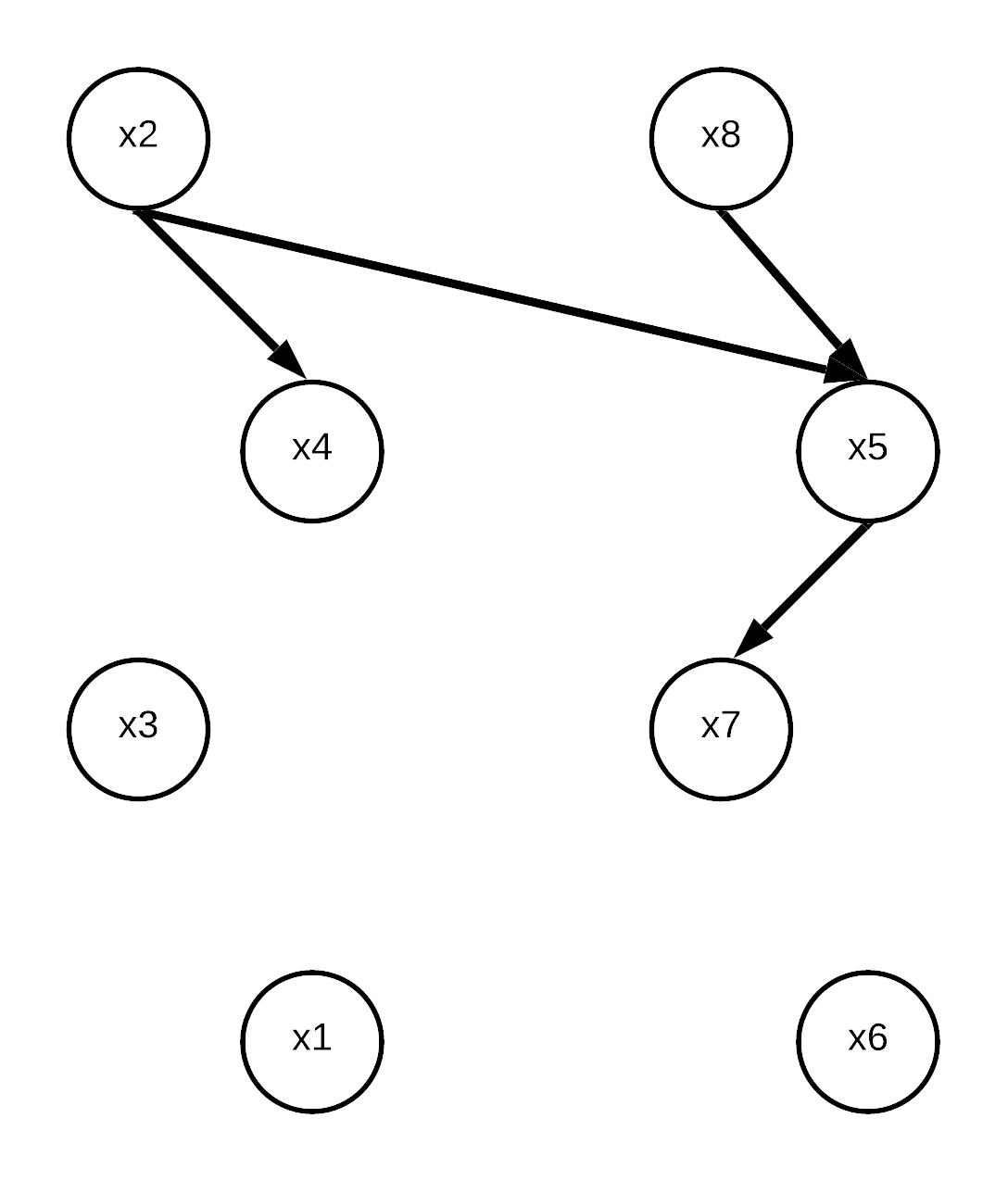}%
}

\subfloat[The Ground Truth Causal Diagram]{%
  \includegraphics[clip,width=0.7\columnwidth,height=6cm]{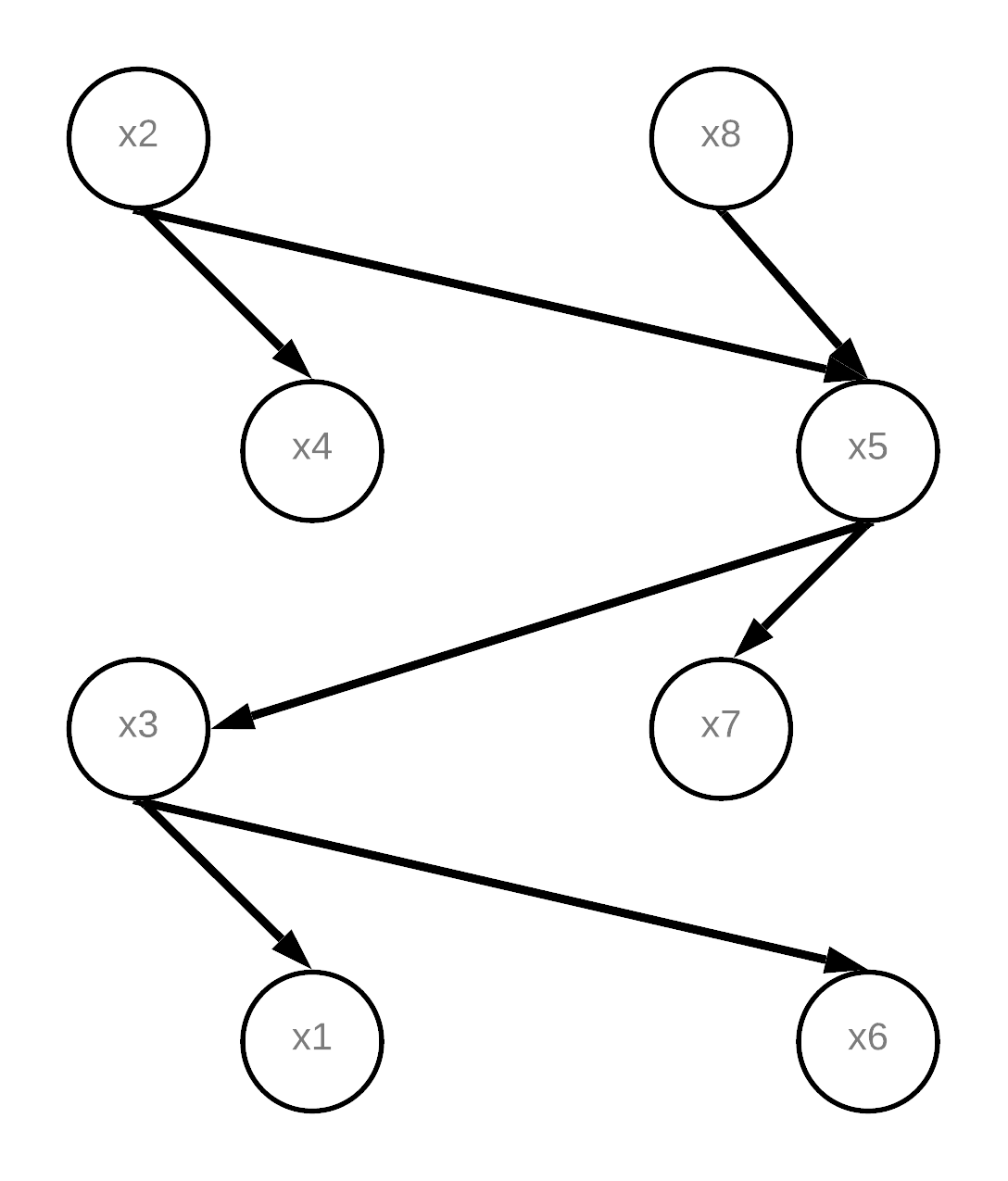}%
}
\caption{Causal Diagram Comparison between GC-ESN with 10\% Missing Data Filling using GPR and the Ground Truth.}
\label{fig:7}
\end{figure}

Figure \ref{fig:7} shows the causal relations recovered from our proposed GC-ESN system (a) side-by-side with its corresponding ground truth causality diagram (b). On the other hand, table \ref{tab:mcc_score} summarizes the MCC index of the proposed system (GPR + ESN 10\% Missing) against the other algorithms. 

\begin{table}[!htb]
    \caption{MCC Score Results}
    \centering
    \resizebox{\columnwidth}{!}{
\begin{tabular}{|c|c|c|c|c|c|}
\hline
    & \begin{tabular}[c]{@{}c@{}}GC-ESN \\ 0\% Missing\end{tabular} & \begin{tabular}[c]{@{}c@{}}GPR + GC-ESN \\ 10\% Missing\end{tabular} & \begin{tabular}[c]{@{}c@{}}GPR + SLARAC \\ 10\% Missing\end{tabular} & \begin{tabular}[c]{@{}c@{}}GPR + MVGC \\ 10\% Missing\end{tabular} & \begin{tabular}[c]{@{}c@{}}SEM \\ 10\% Missing\end{tabular} \\ \hline
TP  & 4                                                             & 4                                                                    & 7                                                                    & 7                                                            & 0                                                           \\ \hline
FP  & 13                                                            & 15                                                                   & 47                                                                   & 48                                                           & 13                                                          \\ \hline
TN  & 45                                                            & 43                                                                   & 10                                                                    & 9                                                            & 44                                                          \\ \hline
FN  & 2                                                             & 2                                                                    & 0                                                                    & 0                                                            & 7                                                           \\ \hline
MCC & 0.29                                                       & 0.26                                                               & 0.15                                                              & 0.14                                                      & -0.18                                                    \\ \hline
\end{tabular}
}
\label{tab:mcc_score}
\end{table}


\noindent{As} shown in Table. \ref{tab:mcc_score}, our system was capable of recovering four of the seven causal relations by applying thresholds that yield the highest F1 scores, and despite the 10 \% missing entries, our proposed system still achieved an MCC index (Table \ref{tab:mcc_score}) of 0.31 which is slightly lower (by 0.03) than the MCC index obtained with the original data (using ESN causal learning). This indicates that the GPR filling process we adapted inside our system is very effective and did restore a satisfactory amount of information comparable to that of the original data.
Furthermore, compared to the remaining algorithms on missing data, the proposed GC-ESN estimator achieved the highest MCC score. This indicates that our proposed system is capable of offering more precise and reliable causal links suggestions than the other systems.

\begin{figure}[!htb]
    \centering
    \includegraphics[width=0.5\textwidth,height=0.3\textheight,keepaspectratio]{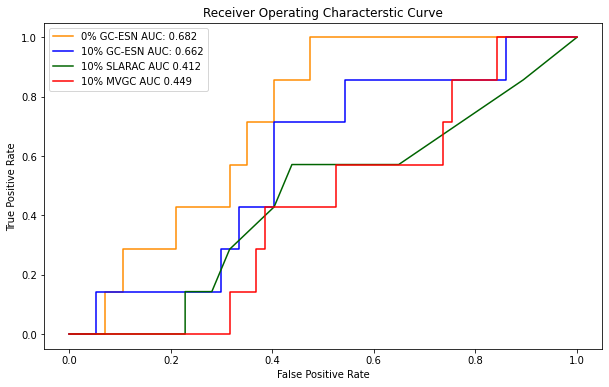}
    \caption{ROC Curve for Different Missing Percentage GC-ESN and other estimators. The proposed system is able to outperform the rest of the estimators by a noticeable margin in their AUC scores.}
    \label{fig:8}
\end{figure}




The results of the ROC curves (shown in Fig. \ref{fig:8}) further solidify our claims as our proposed GC-ESN system reports a significantly larger AUC score than the rest(other than ESN on the original data). 
Moreover, by varying the threshold values, GC-ESN has the highest true positive rate, proving that among the three estimators, GC-ESN exhibits the best performance. 



\section{Conclusion} 

We have proposed a system 
capable of performing causal discovery for sparsely sampled multivariate time series data. The system consists of two parts: (1) Data filling with Gaussian Process Regression, and (2) causal learning with an Echo State Network. The proposed system is evaluated on the Tennessee Eastman (TE) process dataset with 10 percent missing entries. In order for us to evaluate the performance, the proposed system is compared and shown to out perform several other methods including Structural Expectation Maximization (SEM), Subsampled Linear Auto-Regression Absolute Coefficients (SLARAC), and Multivariate Granger Causality (MVGC).\\
We also perform an ablation study to evaluate the effectiveness of the GPR in recovering causal information by comparing its results to that of causal discovery using the original (uncontaminated) data, and found that the proposed data filling process is capable of recovering causal relationships reliably and performed only marginally worse had the full original data was used.  This work shows promise in recovering causal relationships from imperfect data better than current SOTA (State Of The Art) methods.

The obtained results show great potential in applying the proposed system in more complicated real world scenarios as it outperforms all other methods by a comfortable margin in both AUC scores and MCC indices.\\
\noindent{That} being said, the proposed system still fall short in comparison to a human subject-expert in identifying causal relationships;
as such, causal discovery remains an ongoing research topic.

\section*{Acknowledgments}
We would like to thank the Ontario Centres of Excellence (OCE),   Natural  Sciences  and  Engineering  Research Council (NSERC) and ATS Automation Tooling Systems Inc., for supporting this research work.

\bibliographystyle{IEEEtran}  
\bibliography{references.bib}
\end{document}